\documentclass{sig-alternate}

\usepackage{amssymb,amsmath,dsfont}
\usepackage{bm}
\usepackage{algorithm, algorithmic}
\usepackage{graphicx}
\usepackage{subfigure,comment}
\usepackage{hyperref,enumitem}
\usepackage{multirow,tabu}
\usepackage[table]{xcolor}
\usepackage{color}

\newcommand{\commentout}[1]{}

\newcommand{\tab}{\hspace*{2em}}

\DeclareMathOperator*{\argmax}{arg\,max}

\begin{document}
%
% --- Author Metadata here ---
\conferenceinfo{WOODSTOCK}{'97 El Paso, Texas USA}
%\CopyrightYear{2007} % Allows default copyright year (20XX) to be over-ridden - IF NEED BE.
%\crdata{0-12345-67-8/90/01}  % Allows default copyright data (0-89791-88-6/97/05) to be over-ridden - IF NEED BE.
% --- End of Author Metadata ---

\title{MCODE: Multivariate Conditional Outlier Detection}
%%%% \subtitle{[Extended Abstract]}

\numberofauthors{2} %  in this sample file, there are a *total*
\author{
% 1st. author
\alignauthor
Charmgil Hong\\
       \affaddr{Department of Computer Science}\\
       \affaddr{University of Pittsburgh}\\
       \affaddr{Pittsburgh, PA 15213}\\
       \email{charmgil@cs.pitt.edu}
% 2nd. author
\alignauthor
Milos Hauskrecht\\
       \affaddr{Department of Computer Science}\\
       \affaddr{University of Pittsburgh}\\
       \affaddr{Pittsburgh, PA 15213}\\
       \email{milos@cs.pitt.edu}
}
% Just remember to make sure that the TOTAL number of authors
% is the number that will appear on the first page PLUS the
% number that will appear in the \additionalauthors section.

\maketitle
\begin{abstract}
Outlier detection aims to identify unusual data instances that deviate from expected patterns. The outlier detection is particularly challenging when outliers are context dependent and when they are defined by unusual combinations of multiple outcome variable values. In this paper, we develop and study a new conditional outlier detection approach for multivariate outcome spaces that works by (1) transforming the conditional detection to the outlier detection problem in a new (unconditional) space and (2) defining outlier scores by analyzing the data in the new space. Our approach relies on the classifier chain decomposition of the multi-dimensional classification problem that lets us transform the output space into a probability vector, one probability for each dimension of the output space. Outlier scores applied to these transformed vectors are then used to detect the outliers. Experiments on multiple multi-dimensional classification problems with the different outlier injection rates show that our methodology is robust and able to successfully identify outliers when outliers are either sparse (manifested in one or very few dimensions) or dense (affecting multiple dimensions).
\end{abstract}

%A category including the fourth, optional field follows...
%%%% \category{D.2.8}{Software Engineering}{Metrics}[complexity measures, performance measures]


% INTRODUCTION
\section{Introduction}
\label{sec:intro}
Outlier detection is one of the most active topics of research in data mining and statistics. 
%%% with a long history of research and may useful applications. 
The objective of outlier (or anomaly) detection is to find unusual
data instances in the dataset. Outlier detection can be extremely useful for identifying atypical data or behaviors, unusual
outcomes, or erroneous readings and annotations. It is often used as a primary data preprocessing step that helps to remove the noisy or irrelevant signals in a dataset \cite{Hodge:2004:SOD,Liu:2004:CCE}. But most of the time it is utilized to identify interesting (rare) patterns in data that may be associated with either adverse or beneficial events, as in novelty detection \cite{Markou:2003:SP,Pimentel:2014:SP}, fraud identification \cite{Fawcett:1997,Bolton:2002,Wang:2010:ICICTA}, network intrusion surveillance \cite{garcia:2009,Tan:2002,Zhang:2010:IET}, disease outbreak detection \cite{Wong:2003:ICML}, and clinical monitoring and alerting \cite{Hauskrecht:2013}.

Despite huge progress in outlier detection methodologies, the majority of existing outlier detection methods aim to detect unconditional outliers that are identified over the joint space of all data attributes. 
However, these methods are not suitable for many practical problems in which we want to identify unusual (or out of ordinary) responses (labels) associated with data objects. In such a case, outliers depend on the context or properties of the data objects we consider. The application of unconditional methods here may easily lead to both false positives and false negatives detections.  Let us consider, for example, an image annotation (labeling) problem, in which we want to detect erroneous image \textit{tags}. Suppose we applied an unconditional outlier detection approach to this problem. In such a case, images with rare \textit{subjects}, even if their annotations are correct, would be detected due to the scarcity of the subjects in the dataset. Similarly, assume a patient with a rare \textit{disease}. Even though the patient's diagnoses are correct for the manifested symptoms, unconditional outlier detection would incorrectly mark the case as an outlier due to the disease rarity. On the other hand, assume an unusual image, say of some modern painting, is assigned a label that is frequent across the database of images, but incorrect for that specific image or style.  In such a case, label itself is not an outlier when considered without a context but it becomes one when a proper context is considered. Similarly, a moderately high medication dose may look frequent with respect to the patient population that includes both adults and children, but it may become abnormal when considering only children.

The differences between unconditional and conditional outlier detection become apparent when both problems are expressed probabilistically. In conditional outlier detection we seek instances that fall into a low probability region of:
\begin{align}
P({\bf y}|{\bf x}) = P({\bf y},{\bf x})/ P({\bf x})  \notag
\end{align}
where ${\bf y}$ is response (outcome) vector and ${\bf x}$ a data object defining the context. In contrast to this, the unconditional outlier detection approach seeks instances in low probability regions of 
$P({\bf y},{\bf x})$ or $P({\bf y})$.  

The focus of this paper is on the development of conditional outlier detection methodologies in which data objects are associated with multivariate (possibly high dimensional) binary outputs (responses) and our goal is to identify irregularities or rare patterns in these responses. Typically the multivariate binary outputs correspond to label spaces.  Examples of problems that fall in this category are identification of unusual labelings of images, unusual keywords assigned to documents, or incorrect diagnoses associated with the patient case, etc. The conditional outlier detection is particularly challenging in these settings: both context and interdependences in response patterns should be considered when detecting the outliers. 

The approach we propose in this work builds upon the probabilistic classifier chain model \cite{Read:2009:ECML,Boutell:2004:PR,batal:2013:CIKM,Hong:2015:SDM} for multidimensional prediction problems. The model represents  posterior probability of $P({\bf y}|{\bf x})$ by decomposing it into the product of univariate probabilistic predictors $P(y_i|\mathbf{x},\mathbf{y}_{\boldsymbol{\pi}(i)})$,
%$p(y_i|{\bf x}, {\bf y_{\sim i}'})$,
one for each output variable $y_i$, that depend on ${\bf x}$ and values of some other variables in ${\bf y}$, denoted as $\mathbf{y}_{\boldsymbol{\pi}(i)}$.
%${\bf y_{\sim i}'}$. 
These univariate models can be represented and learned using a variety of classic discriminative methods. Briefly, each of the terms of the product represents a probability of observing one dimension of the output space. While one can always calculate the product of these terms to express the full posterior $P({\bf y}|{\bf x})$ (via the chain rule), our approach treats all terms (in the vector form) as a new representation of the output space that accounts for both the context-output and output-output dependences. Our assumption is that the different outlier methods and outlier scores can be successfully defined in this new space. The reason for keeping the terms separate is twofold. First, the errors due to various model estimation procedures are not combined together into one statistic which can make the detection of true irregularities (outliers) hard especially for high dimensional ${\bf y}$. Second, the decomposition lets us adapt the detection procedure to the different types of outliers. For example, when outlier instances are expected to effect only one or just a few dimensions of the output space, the outlier scoring on the new space may focus on the different statistic derived from individual terms as opposed to statistic one would need when outliers are dense and effect many different outputs. For example, when considering the image labeling one may assume the process of generating outliers is random and rare (e.g. in the image labeling a label is randomly added or omitted) and hence a chance seeing outliers in multiple dimensions of ${\bf y}$ is unlikely. On the other, when outliers are expressed over many dimensions (such as in network attacks) the outliers affect many dimensions of the output space. Keeping the space decomposed but still covering key contextual and output dependences helps us to detect more effectively outliers in these different settings. 

We propose and test the different outlier criteria defined upon the new output space that captures context-output and output-output dependences. The experiments are conducted on a number of multi-dimensional classification datasets with the different outlier processes injecting the errors into the output spaces. We demonstrate that our methodology is robust and able to detect outliers when the outlier signal is both sparse (manifested in one or very few output dimensions) and dense (affecting multiple dimensions).

The rest of the paper is organized as follows. 
Section \ref{sec:prob} formally defines the multivariate conditional outlier detection problem we are investigating. 
Section \ref{sec:related} reviews the related research work.
Section \ref{sec:approach} describes the new outlier detection approach.
Section \ref{sec:experiments} presents the experimental results and evaluations. Section \ref{sec:concl} concludes the paper.

%%% end commentout

% PROBLEM DEFINITION
\section{Problem Definition}
\label{sec:prob}

This section provides the formal definitions and notation of the \textit{multivariate conditional outlier detection} problem addressed and researched in this paper. In particular, we consider a special case of the multivariate conditional outlier detection problem where each data instance is associated with $d$ discrete-valued response variables $\mathbf{Y}=(Y_1, ..., Y_d)$. 
We are given training data $D_\textit{train} = \{ \mathbf{x}^{(n)}, \mathbf{y}^{(n)} \}_{n=1}^N$, 
where each observation (context) $\mathbf{x}^{(n)}=(x_1^{(n)}, ..., x_m^{(n)})$ 
is associated with $d$ response (output) variables $\mathbf{y}^{(n)}=(y_1^{(n)}, ..., y_d^{(n)})$.
Our goal is to identify unusual responses in the data that reside in (unseen) testing data $D_\textit{test} = \{ \mathbf{x}^{(n)}, \mathbf{y}^{(n)} \}_{n=N+1}^{N+N'}$.

The fundamental challenges for building multivariate conditional outlier detection model are: \textit{how to build an accurate model representing the dependency of response variables ${\bf y}$ on context variables ${\bf x}$, and mutual dependences among response variables}. We approach this problem by modeling $P(\mathbf{Y|X})$. However, this representation is exponential in the dimensionality of the output space $d$; hence, one of the key questions is how to reduce the complexity of this model. 

\vspace{.5em}
\begin{scriptsize}
\noindent
{\textbf{Notation:} For notational convenience, we will omit the index superscript $^{(n)}$ when it is not necessary. We may also abbreviate the expressions by omitting variable names; e.g., $P(Y_1\!=\!y_1, ..., Y_d\!=\!y_d|\mathbf{X\!=\!x}) = P(y_1, ..., y_d|\mathbf{x})$.}
\end{scriptsize}

% RELATED RESEARCH
\section{Related Research}
\label{sec:related}
%The work in this paper draws and builds upon results of investigations in outlier detection and multidimensional classification. We review each of these directions next, starting with the review of relevant outlier detection \cite{Chandola:2009:ACMCS} 

%\subsection{Outlier detection}

Outlier detection \cite{Chandola:2009:ACMCS,Filzmoser:2004:TUW-114975,Markou:2003:SP,kriegel:2010:sdm} has been studied extensively by data mining and statistics communities. Accordingly, a variety of approaches have been proposed and applied to identify outliers in data and data streams. %In the following we briefly review some of the outlier detection methods that are closely related to our approach.
While outlier detection studies have been conducted by a wide range of communities, the concept is ill-defined, and there is no general consensus on what the definition of outlier is. Probably the most referred definition has been given by Hawkins \cite{hawkins:1980:book}: 
\textit{``An outlier is an observation which deviates so much from the other 
observations as to arouse suspicions that it was generated by a different 
mechanism.''}
Given this rather broad definition, various methods were proposed to find the most deviating instances in a multivariate dataset. 
The methods can be roughly divided into five groups: depth-based approaches, %deviation-based approaches,
distance-based approaches, density-based approaches, and high-dimensional approaches. 

Depth-based approaches assume that outliers are at the fringe of the response space and normal response
are close or in the center of the space. The typical algorithms in this class include
Exploratory Data Analysis \cite{Tukey1977}, Isodepth
\cite{Ruts:1996:CDC:255810.255825}, and Fast Depth Contours 
\cite{Johnson98fastcomputation}. These methods define the depth of
the data $k$ by gradually removing data from convex hulls and data
samples with small depth are reported as outliers. A related method is the One-Class Support Vector Machine \cite{scholkopf:1999:NIPS} which assumes all the training data belong to one class. The resultant decision boundary then defines the region of normal data, whereas the data lie across the boundary are considered as outliers. 

\begin{comment}
Deviation-based approaches assume the outliers are the
outmost data samples from the set, and they attempt to remove the outliers by minimizing the variance. One of the well-known algorithms in this group is the Linear Method for Deviation Detection \cite{Arning96alinear}. Compared with depth-based approaches, the deviation-based approaches do not need to decide on the depth
$k$, and they do not need complicated contour generation process.
\end{comment}

Density-based approaches assume that the density around
a normal data example is similar to the density around its neighbors. Local outlier detection \cite{breunig:2000,papadimitriou:2003:ICDE,jin2001mining,Zhang:1996:BED:235968.233324}, isolation methods \cite{Tang:2002:EEO:646420.693665} are common methods. Compared with the other approaches, density-based approaches are more locally sensitive and tend to achieve better accuracy. A typical representative is a Local Outer Factor (LOF) \cite{breunig:2000}, which is a relative density score estimated by an extended $k$-nearest neighbor approach. LOF indicates the unusualness of an instance, and can be used as an outlier index. This density-based approach has shown good performance in many applications and influenced several subsequent works in the literature \cite{papadimitriou:2003:ICDE,jin2001mining,Zhang:1996:BED:235968.233324}.

Distance-based approaches assume that normal data examples come from
dense neighborhoods, while outliers correspond to isolated points. The typical method is \cite{Rousseeuw:1990} which is one of the early outlier detection methods, that is still used in many applications. The method gives an outlier score to each instance using a robust variant of the Mahalanobis distance \cite{Rousseeuw:1984}, which measures the distance between each instance to the main body of data distribution, such that the instances located far from the rest instances can be identified as outliers. Other methods that fall in this category include Knorr's unified approach \cite{Knorr97aunified}, linearization method
\cite{Angiulli:2002:FOD:645806.670167}, randomized pruning method
\cite{Bay:2003:MDO:956750.956758}, resolution based method
\cite{Fan_anonparametric}, etc. The limitation of the distance-based methods is that they suffer from the curse of
dimensionality problem. The number of parameters in those models
will increase quadratically in the number of dimensions, which makes
them less suitable for high dimensional data.

In the high-dimensional space, one of the greatest challenges is that %the curse of dimensionality. T
the data samples are so sparse 
%in the high dimensional space 
 and there is no meaningful neighborhood in such space. High-dimensional 
approaches are proposed to handle such extreme cases. The typical methods in this class either 
adopt an invariant distance measurement, such as, the angle based outlier detection 
\cite{Kriegel:2008:AOD:1401890.1401946}, or project the data to a lower dimensional subspace, 
such as, grid based subspace outlier detection \cite{Aggarwal:2001:ODH:376284.375668}, 
sufficient dimensionality reduction \cite{Globerson:2003:UAI}, Bayes Exponential Family PCA \cite{Mohamed:2008:NIPS}, 
Sparse PCA \cite{zou:2006:sparse}. More recent methods use Gaussian processes to help matrix 
factorization \cite{lawrence:2009}, explore the structure between independent data \cite{huang:2012:ejs}.

The vast majority of existing outlier detection methods attempts to solve 
the ``unconditional'' outlier detection problem, where data instances 
are compared and analyzed across all attributes.
On the other hand, an increasingly popular approach in recent years is the conditional (or contextual) outlier detection that attempts to identify outliers in a subset of response variables given the values of context variables. 
While several approaches \cite{Song:2007:TKDE,Hauskrecht:2013,valko:2011:icdm} have been proposed to this extent, Song et al. \cite{Song:2007:TKDE} proposed a model-based conditional outlier detection method, that uses a generative data representation to capture the conditional relations between context and response variables, and considers the instances that deviate from this representation as outliers.

Although our proposed solution shares some similarities with Song et al. \cite{Song:2007:TKDE}, there are significant differences: 
\begin{enumerate}[leftmargin=*]
    \setlength\itemsep{-0.1em}
    \item[(1)] To model the underlying data representation, our approach uses a multi-dimensional learning approach that directly learns the conditional probability distribution (a discriminative model); 
On the other hand, \cite{Song:2007:TKDE} uses the Gaussian mixture models to learn the joint distribution $P(\mathbf{x})$ and $P(\mathbf{y})$ separately, and the conditional properties are modeled through a probabilistic mapping function.
    \item[(2)] The parameter learning in our approach exploits the chain decomposition \cite{Read:2009:ECML}, which reduces the multivariate conditional modeling to learning of $d$ classification functions, that makes the method scalable to large data;
However, learning of GMMs in \cite{Song:2007:TKDE} requires  expensive Expectation-Maximization steps, which limits its scalability. 
    \item[(3)] In outlier detection on testing instances, our approach estimates and utilizes the piecewise posterior probability of individual responses $P(y_i|\mathbf{x})$, which not only improves the outlier detection performance to a significant extent, but also makes the method sensitive to low-dimensional outliers (sparse outliers); 
While the GMMs used in \cite{Song:2007:TKDE} are only able to compute the conditional joint probability $P(\mathbf{y}|\mathbf{x})$ (estimating $P(y_i|\mathbf{x})$ computationally infeasible).
\end{enumerate}

%\subsection{Multidimensional prediction}
%ADD CONTENT

% OUR APPROACH
\section{MCODE Model}
\label{sec:approach}

This section describes MCODE, our multivariate conditional outlier detection approach. Briefly, we present a \textit{model-based} outlier detection technique that learns a data model from a training dataset, which is assumed to be outlier-free (or the effect of outliers in modeling is assumed negligible; note that the same assumption is used in \cite{Song:2007:TKDE}), and then uses the model to detect outliers from unseen data, which may include outliers. Accordingly, the proposed approach consists of the following two phases: (1) We first build a probabilistic multivariate conditional model from the training data. (2) The model, when it is applied to different data instances, is used to estimate \textit{outlier scores} that measure how the new data patterns are likely or unlikely based on the trained model. Section \ref{subsec:approach_model} and \ref{subsec:approach_score} describe these two phases in more detail.

\subsection{Conditional Probabilistic Models of Multivariate Outputs}
\label{subsec:approach_model}

Our outlier detection approach summarizes the data by a model which is then used for outlier detection. So our objective first step is to build (from data) an accurate probabilistic model relating context variables $\mathbf{X} = (X_1,...,X_m)$ defining the different data objects and output variables $\mathbf{Y} = (Y_1,...,Y_d)$ defining the response. More specifically, we want to learn an accurate predictive probabilistic model $P(\mathbf{Y}|\mathbf{X})$.

The problem of learning $P(\mathbf{Y}|\mathbf{X})$ from data has been studied extensively in context of \textit{multi-dimensional learning} (MDL) \cite{Gaag:2006:PGM,Zhang:2013} were the goal is to learn $P(\mathbf{Y}|\mathbf{X})$ and use it to support multivariate classification tasks, that will be able to automatically assign tags to new images \cite{Boutell:2004:PR,Qi:2007}; 
keywords or topics to text documents \cite{Kazawa:2005, Zhang:2006}; different functions to genes \cite{Clare:2001:PKDD,Zhang:2006},  
and/or diseases to patients \cite{Pestian:2007}. The assignment task corresponds to finding the maximum a posteriori (MAP) assignment of response variables:
\begin{align}
\mathbf{y}^* &= \argmax_\mathbf{y} P(\mathbf{Y}=\mathbf{y}|\mathbf{X}=\mathbf{x})\\
		&= \argmax_{y_1,...,y_d} P(Y_1=y_1,...,Y_d=y_d|\mathbf{X}=\mathbf{x})
\end{align} 

However, we note that for the purposes of conditional outlier detection, we are not interested in using the model to find the optimal assignment, instead we are interested in assessing how likely the observed context-output assignment is.    

A key challenge in learning $P(\mathbf{Y}|\mathbf{X})$ is that (1) $\mathbf{X}$ can be complex high dimensional space defined by a mixture of discrete and continuous context variables, (2) the number of possible assignments of values to output variables is exponential in $d$. While many different machine learning solutions that address the first problem exist, for example, various discriminative classification techniques enhanced with feature regularization, the second problem is equally important and it is unfeasible to model and learn all possible output assignments independently.  

A simple solution to the output space problem is the \textit{Binary Relevance} (BR) method that assumes all responses $\mathbf{Y}$ are conditionally independent of each other given context $\mathbf{X}$, and learns $d$ functions separately \cite{Boutell:2004:PR,Clare:2001:PKDD}. However, this may not suffice for many real-world modeling tasks where the dependences among the responses hold important information to build an accurate model. 

To introduce the dependences among outputs the \textit{Classifier Chains} (CC) approach \cite{Read:2009:ECML} defines a multi-dimensional model of response variables by decomposing them via the chain rule into a product a univariate conditional models, one model of each variable of the output space. Briefly, CC framework decomposes the multivariate conditional distribution $P(\mathbf{Y|X})$ using a product of the posterior over individual response variables $(Y_1, ..., Y_d)$ as:
\begin{align}
\label{eq:ccf}
P(Y_1,...,Y_d | \mathbf{X}; M) &= \prod_{i=1}^d P(Y_i|\mathbf{X},\mathbf{Y}_{\boldsymbol{\pi}(i,M)}),
\end{align}
where $\mathbf{Y}_{\boldsymbol{\pi}(i,M)}$ denotes the parents of $Y_i$ (or in other words output variables $Y_i$ directly depends on) in a model $M$. 
The framework exploits the decomposable structures of the underlying dependency relations among the response variables $\mathbf{Y}$ which is represented in $M$. 
Note that this representation generalizes the BR, by assuming $M$ does not define any relations among output components (i.e., $\mathbf{Y}_{\boldsymbol{\pi}(i,M)} = \{\}$; an empty set).

% FIGURE: DIAGRAMS FOR EXAMPLE STRUCTURES
\begin{figure}[t]
\centering
	\subfigure[DBR]{\label{fig:dbr-ex}\includegraphics[width=0.11\textwidth]{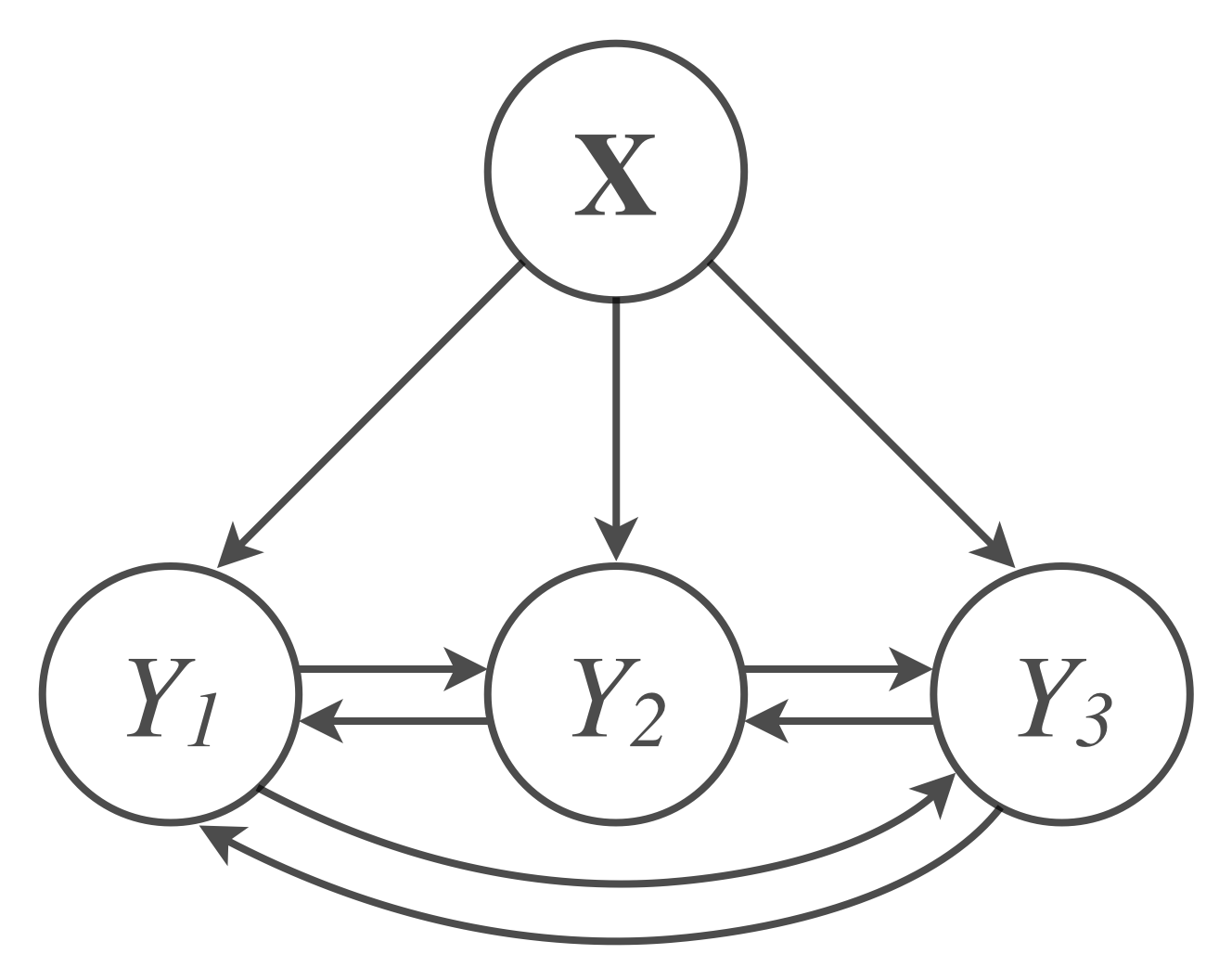}}
	\quad\quad\quad\quad\quad
	\subfigure[BR]{\label{fig:br-ex}\includegraphics[width=0.11\textwidth]{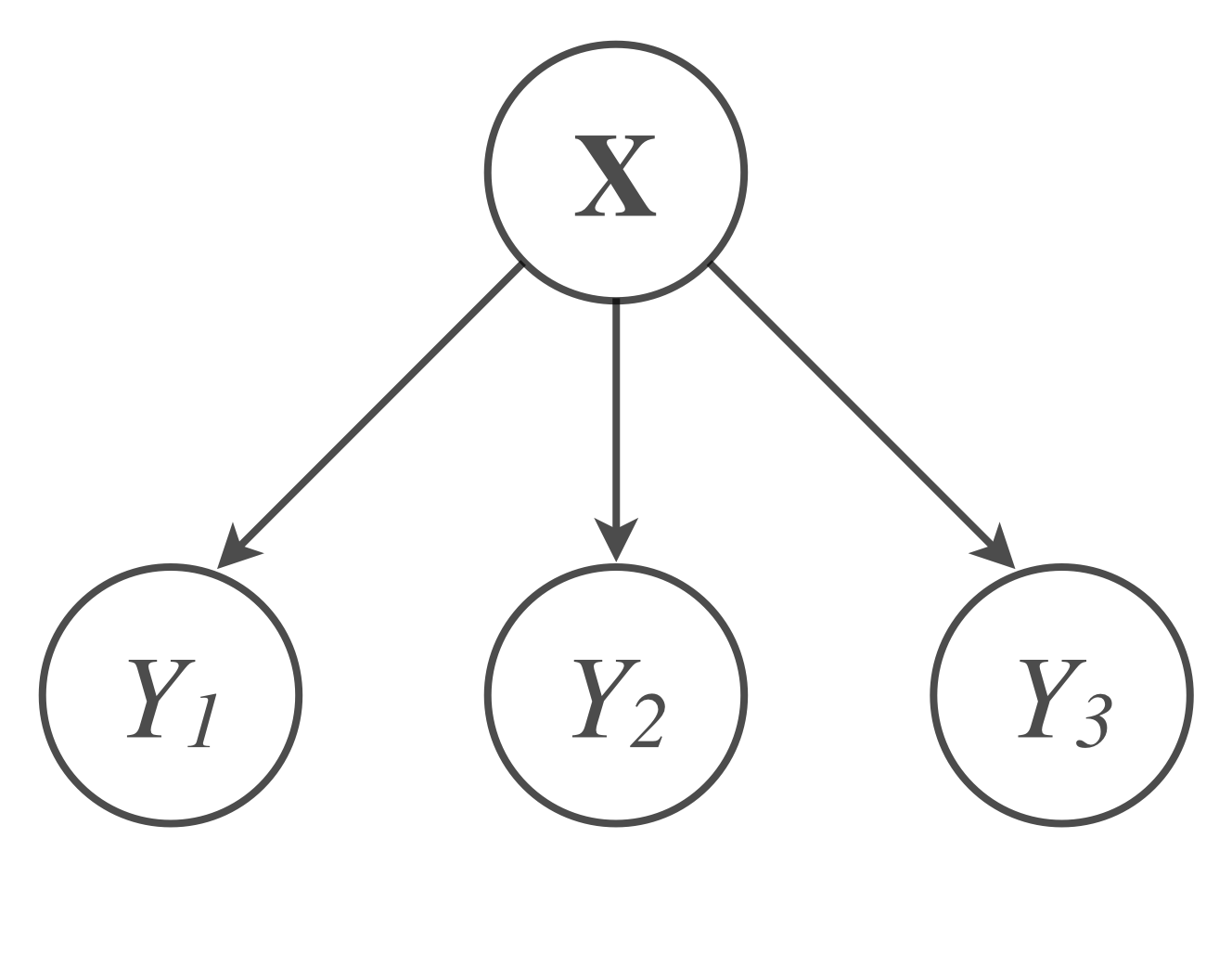}}
	\caption{A comparison of Dependent Binary Relevance (DBR) and Binary Relevance (BR) models in graphical representation ($d = 3$).}
	\label{fig:examples}
\end{figure}

A related decomposition scheme is the \textit{Dependent Binary Relevance} (DBR) model \cite{Montanes:2014:PR}. This model does not adhere to the chain rule decomposing the joint of the output space $P({\bf Y}|{\bf X})$, and it permits circular dependences among output variables. Hence it is best viewed as an approximation of $P({\bf Y}|{\bf X})$, that is,
\begin{align}
\label{eq:ccf-dbr}
P(Y_1,...,Y_d | \mathbf{X}; M) & \sim \prod_{i=1}^d P(Y_i|\mathbf{X},\mathbf{Y}_{\boldsymbol{\pi}(i,M)}),
\end{align}
where 
\begin{align}
%%% \label{eq:ccf-dbr}
\mathbf{Y}_{\boldsymbol{\pi}(i,M)} = \mathbf{Y} \backslash Y_i
				= (Y_1, ..., Y_{i-1}, Y_{i+1}, ..., Y_d)
\end{align}
Figure \ref{fig:examples} shows the graphical representation of DBR and BR when the number of response variables is $3$. Compared with BR, DBR considers the status of all the other response variables in representing data.

We note that our outlier detection approach can be defined and work with many different models that fit the CC like product decomposition \cite{Read:2009:ECML,batal:2013:CIKM,Boutell:2004:PR}.

\subsubsection{Learning}

%\textbf{Parameter learning:} 
The parameter learning of DBR corresponds to specifying the conditional probability distribution (CPD) of each response variable $Y_i$: 
$$P(Y_i|\mathbf{X},\mathbf{Y}_{\boldsymbol{\pi}(i,M)}) = P(Y_i|\mathbf{X},Y_1, ..., Y_{i-1}, Y_{i+1}, ..., Y_d)$$
To represent individual CPDs, we use probabilistic predictive functions, such as logistic regression, support vector machines with probabilistic outputs or the naive Bayes.
In this work, we use logistic regression with $L_2$ regularization.
%\\

%\vspace{-.5em}
%\textbf{Structure learning:} 
Notice that each $Y_i$ is dependent on the rest of the response variables $\mathbf{Y} \backslash Y_i$ and the order of learning CPD does not play an important role in model building.

\begin{comment}
\subsubsection{Probability Estimation}

\end{comment}

\subsubsection{Complexity}

Supposing we use logistic regression as our base probabilistic representation, 
we need $d(m+(d-1)+1) = O(dm+d^2)$ parameters for a DBR model. 
Learning these parameters requires $O(d)$ estimations of $P(Y_i|\mathbf{X},\mathbf{Y}_{\boldsymbol{\pi}(i,M)})$. Hence, the overall complexity of learning a DBR is $O(d)$ times the complexity of learning logistic regression.
%Learning a CCF structure using the above greedy algorithm requires $O(d^2)$ estimations of $P(Y_i|\mathbf{X},\mathbf{Y}_{\boldsymbol{\pi}(i,M)})$. 

\subsection{Identifying Outliers}
\label{subsec:approach_score}

The previous section described how to efficiently learn and represent multivariate data using the DBR \cite{Montanes:2014:PR} model. In this section, we present how to apply the model to unseen testing data and identify multivariate conditional outliers reside in them.

Our objective in the second phase is to estimate the degree of ``outlier-ness'' of unseen data instances using the trained model from the first phase. 
That is, we would like to define effective scoring metrics for a model-based outlier detection. 
An important advantages of DBR towards this objective is that it gives a well-defined model of posterior response probability \cite{Hong:2015:SDM}. Recalling equation (\ref{eq:ccf}), DBR allows an efficient estimation of the pseudo-likelihood $P(\mathbf{Y=y|X=x})$ for any $\mathbf{(x,y)}$ pair. In addition, by exploiting the decomposable structure of the model, we can easily estimate the likelihood of each individual response $y_i$ given its context $\mathbf{x}$; i.e., $P(Y_i=y_i|\mathbf{X=x})$. Namely, given an observation $\mathbf{x}$, how likely/unlikely are individual responses $y_i$ are quantified into a $d$-dimensional vector.

We hypothesize this piecewise posterior probability of individual responses contains crucial information for identifying multivariate conditional outliers, and propose a new outlier detection method along with a set of outlier scoring metrics. 
More specifically, our method first transforms testing data from its original space to the probability space, using the DBR model we obtained from the previous phase.
It then estimates the multivariate outlier scores using the conditional quantities in the new space. 

Although existing model-based conditional outlier detection methods \cite{Song:2007:TKDE} have attempted a similar approach, they are limited in that 
they only use the joint posterior probability $P(\mathbf{y|x})$ by assuming the underlying distribution follows the Gaussian distribution.
As a result, the methods would become less sensitive to the outlying patterns observed in individual dimensions especially when the dimensionality of the data is high; and 
only the patterns deviate from the Gaussian distribution could be detected.
On the other hand, our approach is differentiated in that 
(1) it utilizes the likelihood estimation on each response dimension to identify outliers;
(2) it uses the DBR model (or the CCF models \cite{Hong:2015:SDM}, in general) to represent the data, and does not assume the Gaussian distribution. 
As a result, our proposed approach drives the process of outlier scoring to a more granular level of understanding and utilizing the conditional behaviors in data, and leads to a significant performance improvement in outlier detection.

%TODO: (add) we reduced the conditional outlier detection problem to a problem of multivariate outlier detection where the estimation of observe-decision relations are quantified
%TODO: by doing this, we can capture outlying patterns on each decision dimension

\subsubsection{Outlier Scoring Metrics}
\label{subsubsec:scores}
In this subsection, we describe five outlier scoring metrics that we use in our multivariate conditional outlier detection approach. To recall, our objective is to measure the outlier score of unseen testing data $D_\textit{test} = \{ \mathbf{x}^{(n)}, \mathbf{y}^{(n)} \}_{n=N+1}^{N+N'}$. 
For notational convenience, let us first define a quantity $\boldsymbol{\rho}^{(n)}$ of the $n$-th instance:
\begin{align}
\label{eq:rho}
\boldsymbol{\rho}^{(n)} &= (\rho_1^{(n)}, ..., \rho_d^{(n)}) \\
		&= \left( P(y_1^{(n)}|\mathbf{x}^{(n)},\mathbf{y}_{\boldsymbol{\pi}(i,M)}^{(n)}), ..., P(y_d^{(n)}|\mathbf{x}^{(n)},\mathbf{y}_{\boldsymbol{\pi}(i,M)}^{(n)}) \right),  \notag
\end{align}
where $\mathbf{y}_{\boldsymbol{\pi}(i)}^{(n)} = \mathbf{y}^{(n)}\backslash y_i^{(n)}$, and $M$ denotes the underlying data representation. Using this $d$-dimensional quantity $\boldsymbol{\rho}^{(n)}$, below we define our outlier scoring metrics.

\vspace{1em}\noindent
\textbf{\textit{Score}$_1$: Complementary Probability}

The first outlier scoring metric is a univariate scoring metric that uses the natural interpretation of probability. I.e., the metric takes an instance $(\mathbf{x}^{(n)}, \mathbf{y}^{(n)})$ to estimate the complementary probability based on model $M$. Note that this is a widely used outlier scoring technique \cite{Markou:2003:SP,Song:2007:TKDE}. 

\begin{align}
\textit{Score}_1(\mathbf{x}^{(n)},\mathbf{y}^{(n)}) = 1 - P(\mathbf{y}^{(n)}|\mathbf{x}^{(n)};M)
\end{align}

\vspace{1em}\noindent
\textbf{\textit{Score}$_2$: Robust Distance}

The robust distance \cite{Rousseeuw:1990} measures the deviation between each instance and the main body of distribution, using a robust variant of the Mahalanobis distance \cite{Rousseeuw:1984} method. As a results, the method can maintain a notion of normal data during the process of outlier scoring.
\begin{align}
\textit{Score}_2(\boldsymbol{\rho}^{(n)}) &= \textit{robust.dist}\left( \boldsymbol{\rho}^{(n)} \right)^2 \notag\\
	&= (\boldsymbol{\rho}^{(n)}-\boldsymbol\mu)' C^{-1} (\boldsymbol{\rho}^{(n)}-\boldsymbol\mu),
\end{align}
where $\boldsymbol\mu$ denotes the mean of $\{ \boldsymbol{\rho}^{(n)} \}_{n = N+1}^{N+N'}$, and $C$ is a robust estimation of the covariance matrix \cite{Rousseeuw:1984} on $\{ \boldsymbol{\rho}^{(n)} \}_{n = N+1}^{N+N'}$.

\vspace{1em}\noindent
\textbf{\textit{Score}$_3$: $L_r$ Norms}

For the purpose of multivariate conditional outlier detection, in general, we are more interested in the responses whose likelihood is low. Using $L_r$ norms of $1-\boldsymbol{\rho}^{(n)}$, we increase the contribution of such less likely responses to the outlier score, along with the choice of parameter $r$. 
\begin{align}
\textit{Score}_3(\boldsymbol{\rho}^{(n)}, r) &=  \left|\left| 1 - \boldsymbol{\rho}^{(n)} \hspace{.1em}\right|\right|_r
\end{align}
%In our preliminary experiments, $r=\{1,2,\infty\}$ have shown to be useful in scoring. 
In this paper, we report our results using $r = \infty$ such that only the least likely response ($\max_i (1-\rho_i^{(n)})$) decides the outlier score.

\begin{table}[t]
\centering
\resizebox{.475\textwidth}{!}{
\bgroup
\def\arraystretch{1.5}
    \begin{tabular}{| c || m{1.15in} | l |}
    \hline
    \cellcolor{black!15}~        & \cellcolor{black!15}\textbf{Key Quantity}        & \cellcolor{black!15}\textbf{Metric} \\ \hline\hline
    \cellcolor{black!15}\begin{tabular}[x]{@{}c@{}}\textbf{Univariate}\\\textbf{Metric}\end{tabular}    & Complementary probability        & $Score_1 = 1 - P(\mathbf{y}|\mathbf{x})$ \\ \hline
    \cellcolor{black!15}~ 			& Robust distance 	& $Score_2 = (\boldsymbol{\rho}-\boldsymbol\mu)' M^{-1} (\boldsymbol{\rho}-\boldsymbol\mu)$ \\ \cline{2-3}
    \cellcolor{black!15}~			& $L_r$ norms			& $Score_3 = \left|\left| 1 - \boldsymbol{\rho} \hspace{.1em}\right|\right|_r$ \\ \cline{2-3}
    \cellcolor{black!15}~			& Local outlier factor		& $Score_4 =  \displaystyle\sum_{o \in N_k(\boldsymbol{\rho})}\frac{\textit{lrd}_k (o)}{\textit{lrd}_k (\boldsymbol{\rho})} / | N_k(\boldsymbol{\rho}) |$ \\ [1.5em] \cline{2-3}
    \cellcolor{black!15}\multirow{-4}{*}{\begin{tabular}[x]{@{}c@{}}\textbf{Multivariate}\\\textbf{Metric}\end{tabular}} & One-class SVM score & $Score_5 =\mathbf{w} \cdot \phi(\boldsymbol{\rho}_n) - \sigma$ \\ \hline
    \end{tabular}
\egroup}
\caption{Summary of the outlier scoring metrics. $\boldsymbol{\rho}$ denotes the individual posterior response probability (equation (\ref{eq:rho})).}
\label{table:outliers}
\end{table}

\vspace{1em}\noindent
\textbf{\textit{Score}$_4$: Local Outlier Factor}

Local Outlier Factor (LOF) \cite{breunig:2000} uses a relative density score estimated by an extended $k$-nearest neighbor approach: 
\begin{align}
\label{eq:lof}
\textit{Score}_4(\boldsymbol{\rho}^{(n)},k) &=  \frac{\sum_{o \in N_k(\boldsymbol{\rho}^{(n)})}\frac{\textit{lrd}_k (o)}{\textit{lrd}_k (\boldsymbol{\rho}^{(n)})}}{| N_k(\boldsymbol{\rho}^{(n)}) |},
\end{align}
where $\textit{lrd}_k (\boldsymbol{\rho}^{(n)})$ is the local reachability density of $\boldsymbol{\rho^{(n)}}$ defined as:
\begin{align*}
\textit{lrd}_k (\boldsymbol{\rho}^{(n)}) = \frac{| N_k(\boldsymbol{\rho}^{(n)}) |}{\sum_{o \in N_k(\boldsymbol{\rho}^{(n)})} \max(\textit{k-dist}(o),\textit{dist}(\boldsymbol{\rho}^{(n)},o))}
\end{align*}
which in essence summarizes the density in the neighborhood of $\boldsymbol{\rho}^{(n)}$. 
As a result, LOF estimates the unusualness of an instance in consideration of its local density, compared to the local densities of its neighbors. 
For more technical detail and theoretical discussion, see \cite{breunig:2000}.

\vspace{1em}\noindent
\textbf{\textit{Score}$_5$: One-Class SVM Score}

The last scoring metric is relying on the One-Class Support Vector Machine (OCSVM) \cite{scholkopf:1999:NIPS} technique. For training, OCSVM assumes all the training data belong to one (normal) class and attempts to find the maximum margin hyperplane between data and the origin. 
The following quadratic program formulates the training of OCSVM \cite{scholkopf:1999:NIPS}.
\begin{align}
\min_{\mathbf{w},\xi^{(n)},\sigma} \frac{1}{2} ||w||^2 + \frac{1}{\nu N} \sum_{n=1}^{N} \xi^{(n)} - \sigma
\end{align}
\begin{align}
s.t. \tab& (\mathbf{w} \cdot \phi(\boldsymbol{\rho}^{(n)})) \geq \sigma - \xi^{(n)} &&: \forall n = 1, ..., N\\
&\xi^{(n)} \geq 0 &&: \forall n = 1, ..., N
\end{align}
where slack variables $\xi^{(n)}$ is used with parameter $\nu$ to control the smoothness. 
The resultant decision boundary $f(\boldsymbol{\rho}) = \mathbf{w} \cdot \phi(\boldsymbol{\rho}) - \sigma$ then defines the region of normal data, whereas the instances crossing this boundary are considered as outliers. 
To estimate the outlier score on testing instances, we use the raw output of OCSVM, which represents the relative location of the instances to the decision boundary.

\vspace{1em}
Table \ref{table:outliers} summarizes the outlier scoring metrics discussed in this section. After we obtain the outlier scores for testing data, we once again convert the scores to the percentile rank of the instances. This step allows us to evenly distribute the instances across the full range of the outlier score, and lets us perform a more stable outlier detection.

% EXPERIMENTS
\section{Experimental Results}
\label{sec:experiments}
To validate our approach and demonstrate its effectiveness, we present experimental results on real-world datasets. 
In particular, through this section, we would like to verify (1) whether considering the conditional dependency among response variables improves the performance in outlier detection and (2) whether exploiting the piecewise probabilistic estimation of individual responses is useful in identifying outliers.

The evaluation of the performance in outlier detection, however, is not straightforward. This is due to the unsupervised nature of the task that we do not have knowledge on how outliers exist in a given dataset. Therefore, we make the following assumptions before we design our experiments.
\begin{itemize}[leftmargin=*]
    \setlength\itemsep{-0.1em}
    \item Outliers are the fallouts of a conditional data generation process that assigns to each observation ($\mathbf{x}$) the most probable response ($\mathbf{y}$). Hence, outlying components are not in the observation space but in the response space.
    \item The datasets we use in the experiments may contain a small portion of outliers that, however, do not affect the comparison of methods in general because the fraction is too small to influence our model building process and the resultant data representation.
    \item Although a process of outlying response cannot be known nor modeled, we assume we can create outliers by perturbing the responses in data.
\end{itemize}

Based on these assumptions, we conduct our experiments that consist of two parts. In section \ref{subsec:exp1}, we consider a realistic scenario where a fraction of responses are outlying when they are conditioned on contexts. We compare eight different outlier detection methods on six real-world datasets, and show that our approach produces competitive results. 
In section \ref{subsec:exp2}, on the other hand, we set up a controlled situation where we can adjust the number of incorrect responses can be wrong per outlier. Through the experiments on three real-world datasets, we show our approach is even sensitive to sparse outliers as well as to dense outliers.

\subsection{Experiment 1}
\label{subsec:exp1}

\subsubsection{Data}
\label{subsec:exp1_data}

In the first part of our experiments, we evaluate the general performance of our outlier detection approach. We use six multi-dimensional datasets obtained from multiple domains.\footnote{The datasets are publicly available at \url{http://mulan.sourceforge.net}} These include semantic video/image labeling (\textit{Mediamill} \cite{Snoek:2006:MM}, \textit{Corel5k} \cite{Duygulu:2002:ECCV}), text categorization (\textit{Bibtex} \cite{Katakis:2008:ECML}, \textit{Reuters} \cite{Lewis:2004:JMLR}), clinical patient classification {\textit{Medical} \cite{Pestian:2007}} and biology (\textit{Genbase} \cite{diplaris:2005}). Each dataset consists of \textit{continuous} features, which represents observation (context), and associated \textit{binary} labels, which represents response. 
Table \ref{table:datasets} summarizes the characteristics of the datasets, including dataset size, label cardinality (the average number of labels per instance), distinct label set (the number of distinct class configurations that appear in the data) and data domain.

% table:datasets
\begin{table}[t]
\centering
\resizebox{.45\textwidth}{!}{
	\begin{tabular}{ c  c  c  c  c  c  c }
		\hline
		\textbf{Dataset} &  \textsc{$N$} &  \textsc{$m$} &  \textsc{$d$} &  \textbf{LC} & \textbf{DLS} & \textbf{DM}\\
		\hline\\[-.85em]
		Mediamill & 43,907 & 120 & 101 & 4.38 & 43,905 & Video\\
		Bibtex & 7,395 & 1,836 & 159 & 2.40 & 7,384 & Text\\
		Reuters & 6,000 & 47,236 & 101 & 2.88 & 5,990 & Text\\
		Corel5k & 5,000 & 499 & 374 & 3.52 & 4,999 & Image\\
		Medical & 978 & 1,449 & 45 & 1.24 & 58 & Clinical\\
		Genbase & 662 & 1,185 & 27 & 1.25 & 24 & Biology\\
		\hline
	\end{tabular}}
	\caption{ Datasets characteristics. ($N$: number of instances, $m$: number of features (observation), $d$: number of labels (response), LC: Label cardinality, DLS: distinct label set, DM: domain)}
	\label{table:datasets}
\end{table}

\vspace{0.5em}
\noindent
\textbf{Creating Synthetic Outliers}

In this part of our experiments, we simulate plausible scenarios where responses can be outlying in given contexts, which are found virtually everywhere. 
For example, in semantic video/image labeling (\textit{Mediamill}, \textit{Corel5k}), a video clip or image may have irrelevant tags;  
in clinical diagnosis (\textit{Medical}), a patient may receive an inaccurate diagnosis; and, 
in gene function analysis (\textit{Genbase}), a gene sequence may be associated with wrong functional labels.

To simulate them, we inject outliers into the response space by the following sequence:
\begin{enumerate}[leftmargin=*]
    \setlength\itemsep{-0.1em}
    \item[1:] \textit{Bootstrap testing data with size 5,000 (optional).}
    \item[2:] \textit{Perturb 0.5\% of response variables uniformly at random, with no pre-selection nor prioritization of either instances or response dimensions.}
\end{enumerate}

After a perturbation process, we will have a bootstrapped test dataset with $\leq 0.5 \%$ of outliers. Note that the bootstrap step is optional for smaller sized datasets, on which only few outliers would be injected and, hence, we cannot perform a proper statistical evaluation.

\begin{table*}[t]
\centering
\resizebox{.975\textwidth}{!}{
    \begin{tabular}{r|lccccccccc}
    ~         & 	& \multicolumn{3}{c}{\textit{Multivariate Methods}}  & ~ & \multicolumn{5}{c}{\textit{Multivariate Conditional Outlier Detection}}         \\ \cline{3-5} \cline{7-11}
    \multicolumn{1}{c|}{\multirow{-2}{*}{AUC}} & & \multicolumn{1}{c}{RB} & \multicolumn{1}{c}{LOF} & \multicolumn{1}{c}{OCSVM} & ~ & \multicolumn{1}{c}{MCODE-ComP} & \multicolumn{1}{c}{MCODE-RB} & \multicolumn{1}{c}{MCODE-$L_\infty$} & \multicolumn{1}{c}{MCODE-LOF} & \multicolumn{1}{c}{MCODE-OCSVM} \\ [.2em] \hline 
    Mediamill 	& & 0.734 (0.187)      & 0.820 (0.045) 	& 0.780 (0.031)     & ~ & 0.962 (0.019)        & 0.931 (0.025)      & \textbf{0.974 (0.017)}     & 0.892 (0.040)       & 0.921 (0.022)         \\
    Bibtex    	& & 0.501 (0.049)      & 0.807 (0.035)   	& 0.512 (0.056)     & ~ & 0.839 (0.032)        & \textbf{0.977 (0.088)}      & 0.888 (0.032)     & 0.930 (0.025)       & 0.762 (0.037)         \\
    Reuters   	& & 0.512 (0.051)      & \textbf{1.000 (0.000)}   	& 0.538 (0.054)     & ~ & 0.903 (0.022)        & 0.978 (0.017)      & 0.960 (0.013)     & \textbf{1.000 (0.000)}       & 0.823 (0.034)         \\
    Corel5k     	& & 0.516 (0.059)      & 0.947 (0.031)   	& 0.630 (0.062)     & ~ & 0.828 (0.029)        & 0.525 (0.086)      & 0.868 (0.029)     & \textbf{0.975 (0.018)}       & 0.795 (0.038)         \\
    Medical     	& & 0.516 (0.051)      & \textbf{1.000 (0.000)}   	& 0.562 (0.048)     & ~ & 0.963 (0.013)        & 0.633 (0.216)      & 0.965 (0.014)     & \textbf{1.000 (0.000)}       & 0.936 (0.024)         \\
    Genbase    	& & 0.512 (0.054)      & \textbf{1.000 (0.000)}   	& 0.848 (0.033)     & ~ & 0.986 (0.020)        & 0.975 (0.102)      & 0.986 (0.020)     & \textbf{0.998 (0.006)}       & 0.987 (0.018)         \\ \hline
    Rank             & & 8.00 (0.00)         &  \textbf{2.83 (2.11)}          & 6.83 (0.40)         & ~  & \textbf{3.92 (1.02)}      & \textbf{4.33 (2.34)}          & \textbf{3.08 (1.20)}         & \textbf{2.17 (1.17)}        & \textbf{4.83 (1.44)}       \\ \hline
    \end{tabular}}
	\caption{[Experiment 1] The mean and standard deviation (in parentheses) of the area under the receiver operating characteristic curve (AUC). The best methods (by paired t-test at $\alpha=0.05$) on each dataset are shown in bold. The last row shows the mean and standard deviation in the ranks of the methods (by the Friedman test followed by Holm's step-down procedure at $\alpha=0.05$).}
	\label{table:exp1_auc}
\end{table*}

\begin{figure*}
\centering
	\subfigure[\textit{Mediamill}]{\label{fig:exp1a}\includegraphics[width=0.3\textwidth]{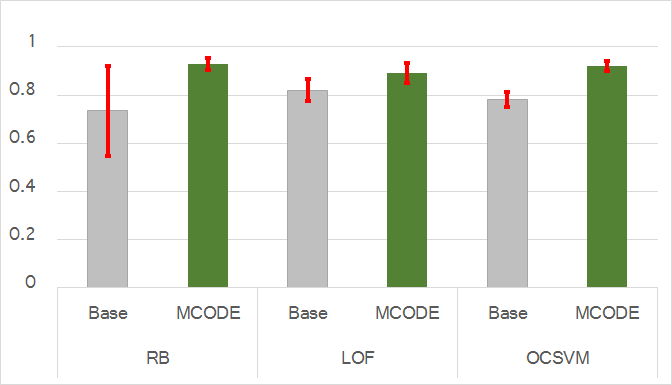}}
	\quad
	\subfigure[\textit{Bibtex}]{\label{fig:exp1b}\includegraphics[width=0.3\textwidth]{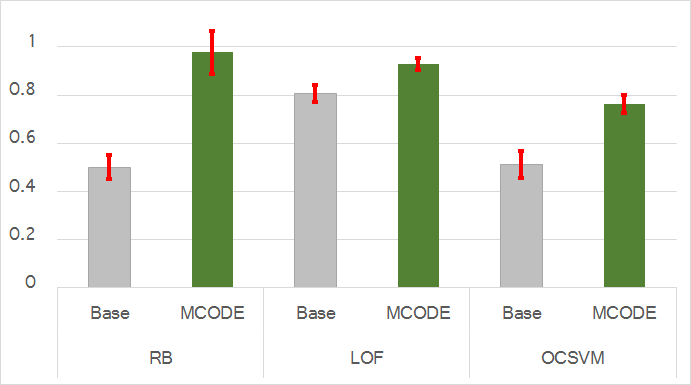}}
	\quad
	\subfigure[\textit{Reuters}]{\label{fig:exp1c}\includegraphics[width=0.3\textwidth]{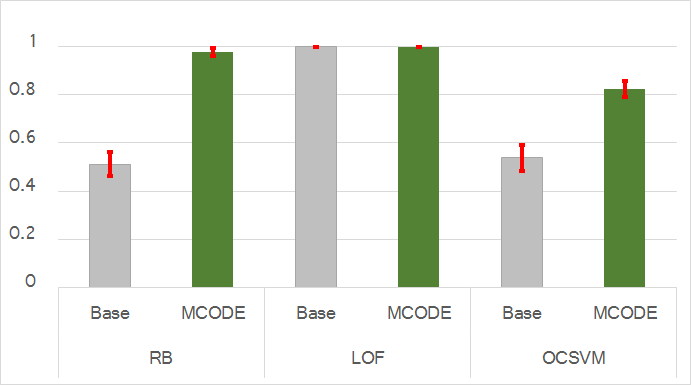}}
	\\
	\subfigure[\textit{Corel5k}]{\label{fig:exp1d}\includegraphics[width=0.3\textwidth]{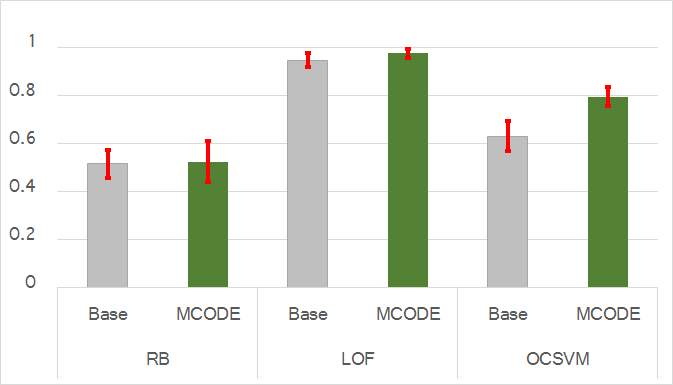}}
	\quad
	\subfigure[\textit{Medical}]{\label{fig:exp1e}\includegraphics[width=0.3\textwidth]{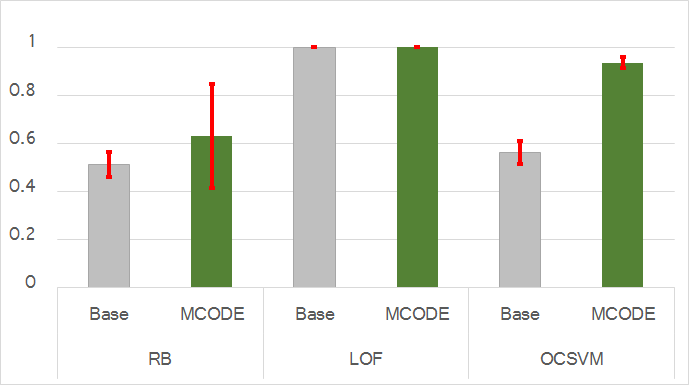}}
	\quad
	\subfigure[\textit{Genbase}]{\label{fig:exp1f}\includegraphics[width=0.3\textwidth]{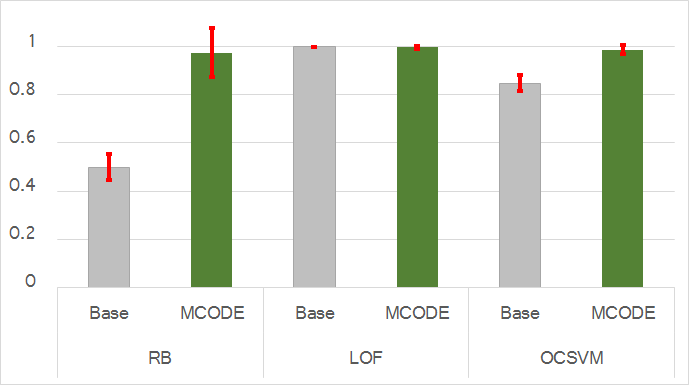}}
	\caption{[Experiment 1] The comparisons of existing multivariate outlier detection methods (gray) and their use in our multivariate conditional outlier detection (MCODE) approach (green) in terms of the area under the receiver operating characteristic curve (AUC). The x-axis indicates different outlier detection methods. The y-axis indicates AUC. The red vertical bars show the standard deviation.}
	\label{fig:exp1_bar}
\end{figure*}

\subsubsection{Methods}
\label{subsec:exp1_methods}

We compare the performance of our approach with other widely used multivariate outlier detection methods, including the \textit{Robust Distance} (RD) \cite{Rousseeuw:1990} approach, \textit{One-class SVM} (OCSVM) \cite{scholkopf:1999:NIPS} and \textit{Local Outlier Factor} (LOF) \cite{breunig:2000}. 
To use these methods, we concatenate each observation and its associated responses into one vector, so that the methods can run over the joint space of all data attributes.
To evaluate our \textit{multivariate conditional outlier detection} (MCODE) approach, we use \textit{Dependent Binary Relevance} (DBR) \cite{Montanes:2014:PR} as the base data model, and apply the five scoring metrics presented in section \ref{subsubsec:scores}. We refer to them with the following identifiers: 
MCODE-ComP uses the complementary probability score; 
MCODE-RD uses the Robust distance \cite{Rousseeuw:1990} score; 
MCODE-$L_\infty$ uses the $L_\infty$ norm score; 
MCODE-OCSVM uses the one-class SVM \cite{scholkopf:1999:NIPS} score; and
MCODE-LOF uses the Local Outlier Factor \cite{breunig:2000} score.

\begin{comment}
\begin{itemize}[leftmargin=*]
    \setlength\itemsep{-0.1em}
    \item MCODE-ComP uses the complementary probability score
    \item MCODE-RD uses the Robust distance \cite{Rousseeuw:1990} score
    \item MCODE-$L_\infty$ uses the $L_\infty$ norm score
    \item MCODE-OCSVM uses the one-class SVM \cite{scholkopf:1999:NIPS} score
    \item MCODE-LOF uses the Local Outlier Factor \cite{breunig:2000} score
\end{itemize}
\end{comment}

For a fair comparison, we fix the following parameters throughout all experiments: 
To train the SVM classifiers for OCSVM and MCODE-OCSVM, we use the radial basis function (RBF) kernel; we set the OCSVM parameter $\nu = 0.01$. 
For LOF and MCODE-LOF, the number of neighbors $k$ is fixed to 30 as used in their original work \cite{breunig:2000}. 
We use $L_2$-penalized logistic regression for DBR; we choose the regularization parameter by cross validation.  

Lastly, recall that OCSVM is a semi-supervised method and, in order to use it as a scoring metric (MCODE-OCSVM), we need to train a classifier which takes the posterior probability of individual responses ($\boldsymbol\rho$) as inputs. Notice that, to avoid overfitting, the data to train OCSVM should be a different subset from the data used to train the DBR model. 
To do this, in all experiments, we use only the half of training data to train DBR, and hold out the rest for the training of OCSVM.

\subsubsection{Metric}
\label{subsec:exp1_models}
We use the \textit{area under the receiver operating characteristic curve} (AUC) to evaluate different methods. AUC is a single number summary of the ROC curve which draws the ratio between true positive rate (TPR) and false positive rate (FPR) by sweeping the threshold over the range of output scores. 
AUC is particularly useful when the optimal decision threshold is unknown. Note that the higher AUC is, the better the performance is.

\begin{figure*}
\centering
	\subfigure[AUC-PR on \textit{Mediamill}]{\label{fig:aucpr1}\includegraphics[width=0.325\textwidth]{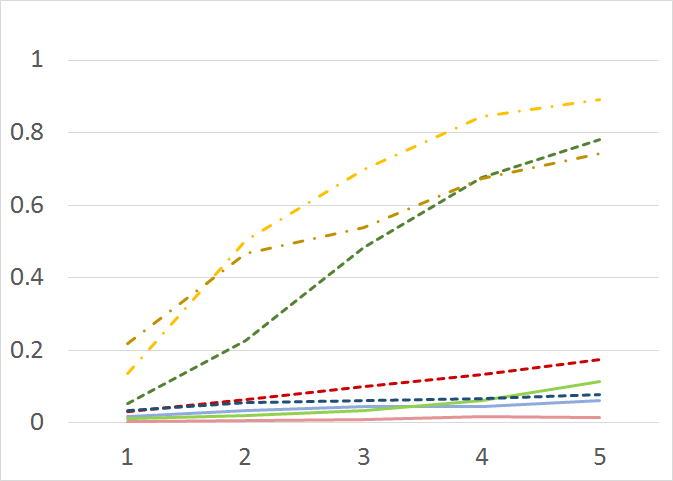}}
	\subfigure[AUC-PR on \textit{Bibtex}]{\label{fig:aucpr2}\includegraphics[width=0.325\textwidth]{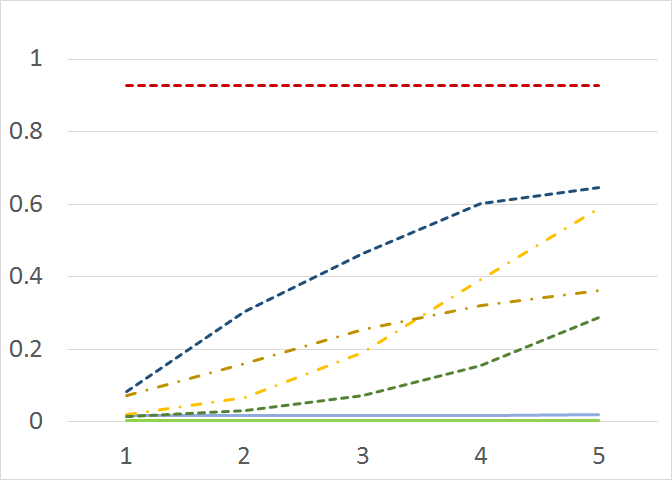}}
	\subfigure[AUC-PR on \textit{Corel5k}]{\label{fig:br-ex3}\includegraphics[width=0.325\textwidth]{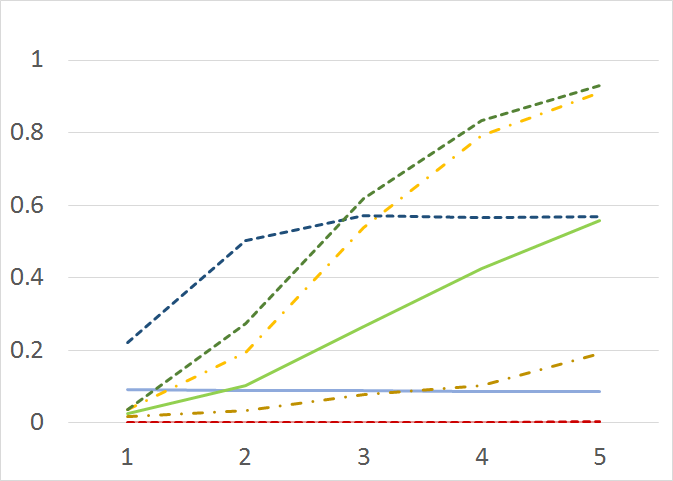}}
	\\
	\subfigure{\includegraphics[width=0.65\textwidth]{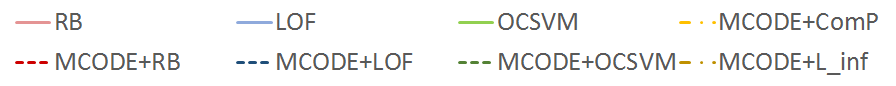}}
	\caption{[Experiment 2] The changes in the area under the precision-recall curve (AUC-PR) according to the different outlier injection rates. The x-axis indicates outlier injection rates. The y-axis indicates AUC-PR.}
	\label{fig:exp2_line}
\end{figure*}

\subsubsection{Results}
\label{subsec:exp1_results}
Table \ref{table:exp1_auc} shows the \textit{area under the receiver operating characteristic curve} (AUC) of the compared methods. We have performed \textit{ten-fold cross validation} with \textit{three repeats} for all of the datasets. The mean and standard deviation (in parentheses) over 30 runs are reported. On each dataset, we mark the best methods and their statistically equivalent methods (by paired t-tests at 0.05 significance level) in bold. The last row shows the mean and standard deviation in the ranks of the methods computed by the Friedman test followed by Holm's step-down procedure with a 0.05 significance level \cite{Demsar:2006, Holm:1979}. Again, the statistically superior methods are marked in bold.

We can see that our approach consistently produces competitive AUC scores. For example, MCODE-LOF outperforms the other methods on four datasets; MCODE-RB and MCODE-$L_\infty$ outperform the other methods on one of the datasets, respectively.
Among the baseline methods, LOF is shown as a close competitor. It produces the best AUCs on three datasets, and results in competitive AUCs on the rest three datasets. We attribute this to the process of its relative density estimation (Equation (\ref{eq:lof})). I.e., the computation of local densities in LOF can be understood as an estimation of likelihood conditioned on local information. 
As a result, LOF can effectively approximate the conditional probability estimation. 
On the other hand, RB and OCSVM do not seem to properly handle the multi-dimensional data. Their unconditional approaches to identify outliers over the joint space of all data attributes do not show much efficacy. This is partially due to the high-dimensionality of the data in

One way to assort the results and analyze the benefits of our approach is to directly compare each baseline and its counterpart in MCODE. Figure \ref{fig:exp1_bar} compares RB, LOF and OCSVM from this perspective. 
The y-axis indicates AUC. The x-axis indicates different methods. 
The results are grouped by the scoring techniques, where the gray bars show the baseline results, and the green bars show that of MCODE. We can see significant improvement from the baseline to MCODE, especially on RB and OCSVM. 
Although, as described above, the performance of LOF is already good as is, by directly working on the conditional probability space, MCODE-LOF even improves the AUC scores.

In summary, the experimental results demonstrate that our MCODE methods, which transforms testing data from its original space to the conditional probability space, actually helps in the identification of outliers and, hence, improves the results.

\subsection{Experiment 2}
\label{subsec:exp2}

\subsubsection{Data}
\label{subsec:exp2_data}

In the second part of our experiments, we would like to test the sensitivity of the methods to the number of outlying dimensions; i.e., we are moving from sparse (each outlying instance has one or very few outlying dimension) to denser (each manifests multiple outlying dimensions) outliers, and test how well each method performs along with this change.

We use three of the multi-dimensional datasets: \textit{Mediamill} \cite{Snoek:2006:MM} (video annotation), 
\textit{Bibtex} \cite{Katakis:2008:ECML} (text categorization) and \textit{Corel5k} \cite{Duygulu:2002:ECCV} (image labeling).
See table \ref{table:datasets} for the characteristics of the datasets. 

\vspace{0.5em}
\noindent
\textbf{Creating Synthetic Outliers}

In this part, we use a rather controlled setting, where we can adjust the number of outlying dimensions. 
Note that this can be a very useful testing protocol in practice, especially for the problems where experts are involved in data labeling (e.g., , making clinical decisions).

To simulate such scenarios, we inject outliers into the response space by the following sequence:
\begin{enumerate}[leftmargin=*]
    \setlength\itemsep{-0.1em}
    \item[1:] \textit{Bootstrap testing data with size 5,000 (optional).}
    \item[2:] \textit{Select 0.5\% of instances uniformly at random.}
    \item[3:] \textit{For each selected instances, select $p$ response dimensions uniformly at random; Perturb the values in the selected dimensions.}
\end{enumerate}

After a perturbation process, we will have a bootstrapped test dataset with exactly $0.5 \%$ of outlier instances, where each outlier has $p$ outlying dimensions.

\subsubsection{Metric}
\label{subsec:exp2_models}

We use the \textit{area under the precision-recall (PR) curve} (AUC-PR). Similar to the AUC score, AUC-PR is the one number summary of the PR curve. While the score is relatively more conservative than AUC, it is useful to depict the sensitivity of methods particularly when the target distribution is imbalanced, as in the outlier detection tasks. 
%The higher AUC-PR is, the better the performance is.

\subsubsection{Results}
\label{subsec:exp2_results}

Figure \ref{fig:exp2_line} shows the AUC-PR of the methods. We have performed \textit{ten-fold cross validation} with \textit{three repeats} for all experiments. 
The y-axis indicates AUC-PR. The x-axis indicates the number of outlying dimensions. 
We use different colors and shapes (solid or dotted) to indicate different methods. 
Simply speaking, the dotted lines show the AUC-PR of MCODE, which are superior in general, whereas the solid lines show that of the baselines.

Intuitively, the smaller the outlying dimension is, the harder the outliers are to be detected. 
Such trends are well captured in the figure \ref{fig:exp2_line}. 
Most methods start from the bottom quarter of the plots, and gradually improve as the number of outlying dimension increases. 
However, we can see the MCODE methods usually start at relatively higher AUC-PRs.
As the number of outlying dimension increases, the differences become more obvious. That is, the AUC-PRs of MCODE grow rapidly, while that of the baseline methods are relatively slower (OCSVM), or seem invariant (RB and LOF) up to this small number of outlying dimensions. 

In summary, this part of our experiments verifies that exploiting the piecewise posterior response probability not only helps to improve the outlier detection performance in general, but also makes the methods more sensitive to the small degree of perturbations.

% CONCLUSION
\section{Conclusions}
\label{sec:concl}

We studied a special case of outlier detection problem where outliers are context dependent and when they are defined by unusual combinations of multiple outcome variable values. 
We reviewed existing outlier detection approaches and multi-dimensional learning methods 
and presented a new conditional outlier detection approach for multivariate outcome space. 
The key motivation of our approach is that we can transform the conditional outlier detection to an unconditional space, and solve the problem more effectively. 
Accordingly, we defined five outlier scoring metrics by analyzing the data in the new space. 
Experiments on two outlier detection settings demonstrate that our approach is not only competitive, but also sensitive to sparse outliers.

%\end{document}  % This is where a 'short' article might terminate

%ACKNOWLEDGMENTS are optional
%\section{Acknowledgments}
%This section is optional

\bibliographystyle{abbrv}
\bibliography{mcode}  % sigproc.bib is the name of the Bibliography in this case
% ACM needs 'a single self-contained file'!
%
% That's all folks!

% Possible appendix? comparison of CLL-loss on BR vs DBR?

\end{document}